\documentclass{llncs}
\usepackage{graphicx} 

\usepackage{algorithm2e}

\usepackage[inline]{enumitem}
\newlist{remarks}{enumerate}{1}
\setlist[remarks]{leftmargin=1.68em,label={\sffamily R$_{\arabic{remarksi}}$}}

\usepackage{booktabs}
\usepackage[hyphens]{url}
\usepackage{xurl}
\usepackage{hyperref}

\usepackage{type1cm}

\usepackage{xcolor}
\definecolor{labelcolor}{cmyk}{0.22,0.10,0.10,0.10}
\definecolor{listbackgroundcolor}{cmyk}{0.10,0.10,0.05,0.05}
\definecolor{listbackgroundcolorlight}{rgb}{0.91,0.92,0.94}
\definecolor{keywordColor}{HTML}{800080}
\definecolor{figPlatBlue}{HTML}{2471A3}
\definecolor{figPlatLight}{HTML}{D6EAF8}
\definecolor{figProxyGreen}{HTML}{1E8449}
\definecolor{figProxyLight}{HTML}{D5F5E3}
\definecolor{figExtGray}{HTML}{616A6B}
\definecolor{figExtLight}{HTML}{EAECEE}

\usepackage{centernot}
\newcommand{\nogoArr}{\textcolor{keywordColor}{\ensuremath{\centernot\longrightarrow}}}

\usepackage[formats]{listings}
\usepackage{multicol}
\lstdefineformat{LangshawFormat}{-/>=\textcolor{keywordColor}{\nogoArr}}
\lstdefinelanguage{Langshaw}
{
  morekeywords={who, what, do, sayso, nono, nogo, key},
  otherkeywords={>,->,-/,:},
  morecomment=[l]{//}
}

\lstdefinelanguage{Peach}
{
  morekeywords={self, systems, agents, protocol, roles, infrastructure, synchronous, proactive, patient, enabled, attempt, AttemptFailed, nogoViolations, nonoViolations, lowSaysos, loadMAS, pending},
  otherkeywords={>,->,-/,:},
  morecomment=[l]{//}
}

\usepackage{standalone}
\usepackage{tikz}
\usetikzlibrary{calc,backgrounds,arrows,shapes}

\usetikzlibrary{positioning}
\usetikzlibrary{arrows,shapes,automata,petri}
\usetikzlibrary{matrix}
\usetikzlibrary{fit}
\usetikzlibrary{decorations}
\usetikzlibrary{decorations.markings}
\usetikzlibrary{decorations.pathreplacing}
\usetikzlibrary{calc}
\usetikzlibrary{graphs}
\usetikzlibrary{quotes}
\usetikzlibrary{arrows.meta}

\tikzstyle{thrust}=[circle,font=\footnotesize\sffamily,fill=colorRepresentationBack,text=colorEntityBack,solid,draw=red, thick,minimum width=12]%

\tikzstyle{message}=[->,-{Classical TikZ Rightarrow[length=1.5mm]},>=stealth']

\usepackage{xspace}
\newcommand{\etal}{{et al.\@\xspace}}

 \usepackage{amssymb}

\usepackage[scr=boondoxo,scrscaled=1.05]{mathalfa}

\newcommand{\fbf}{\textbf}
\newcommand{\fsl}{\textsl}

\newcommand{\fsf}[1]{{\footnotesize{\textsf{#1}}}}

\newcommand{\fn}{\textsf}

\newcommand{\mname}[1]{\fsl{#1}}
\newcommand{\pname}[1]{\fsl{#1}}
\newcommand{\rname}[1]{\textsc{#1}}
\newcommand{\val}[1]{\texttt{#1}}
\newcommand{\paraname}[1]{\fsf{#1}}

\usepackage{fontawesome}

\usepackage{mathpartir}

\DeclareMathAlphabet{\mathcal}{OMS}{cmsy}{m}{n}

\DeclareRobustCommand{\nUmErAL}[1]{#1}

\usepackage[numbers,sort]{natbib}

\usepackage{mdframed}

\newcounter{akccount}

\newcommand{\hide}[1]{}
\newcommand{\mps}[1]{}
\newcommand{\shc}[1]{}
\newcommand{\ai}[1]{}

\title{Strabo: Declarative Specification and Implementation of Agentic Interaction Protocols}
\author{Samuel H. Christie V\inst{1} \and Munindar P. Singh\inst{1} \and Amit K. Chopra\inst{2}}

\date{}

\authorrunning{Christie \emph{et al.}}
\institute{%
North Carolina State University\\
\email{\{schrist, singh\}@ncsu.edu} \and
Lancaster University\\
\email{amit.chopra@lancaster.ac.uk} 
}

\begin{document}
\maketitle

\begin{abstract}

The last few years have witnessed major advances in the modeling and implementation of multiagent systems based on declarative interaction protocols.
Our contribution, Strabo, establishes the relevance of these advances to ongoing industry efforts in Agentic AI.  Specifically, we consider UCP, the \emph{Universal Commerce Protocol}, a recent Google-led effort to standardize e-commerce interactions for AI agents.
Our exercise is in two parts.
One, we model the part of UCP dealing with checkouts as a declarative Langshaw protocol and implement agents using Peach, a programming model for Langshaw.
This part of the exercise brings out the advantages of formal, declarative specifications.
Two, we show that Peach agents can interoperate with UCP agents implemented by Google, thereby establishing the fidelity of our approach with respect to UCP.
Such interoperation enables the incremental introduction of declarative protocols and agents into a conventional setting, indicating a pathway by which EMAS ideas could influence practice without demanding a wholesale update.
\end{abstract}

\section{Introduction}
Interaction protocols model multiagent systems and serve as blueprints for engineering agents. 
Recent work, as exemplified by BSPL \cite{AAMAS-BSPL-11,AAMAS-BSPL-12,IJCAI-21:Tango} and Langshaw \cite{IJCAI-24:Langshaw}, has emphasized formal, declarative approaches for specifying protocols.   
Formal specification enables verifying protocols for correctness before any agents are implemented. 
Declarative specifications better reflect stakeholder intuitions about the meaning of interactions \cite{Computer-98} and, consequently, support flexible interactions between agents.
Programming models that facilitate implementing agents based on declarative protocols \cite{JAAMAS-22:Mandrake,AAMAS-23:Kiko,AAAI-25:Orpheus,AAMAS-26:Peach} support improved code structure and dealing with changing requirements. 
Protocols are conceptual abstractions.
Not specifying them doesn't mean they don't have to be implemented.
It's just that, absent explicit protocols, developers end up encoding the relevant interaction constraints directly in low-level code, which is cumbersome and hides assumptions and errors.  

The Agentic AI paradigm has emerged concurrently with the advances in protocols.  
Its basic value proposition is that LLM-based agents will do all kinds of tasks on behalf of their users.  
Many of these tasks will involve interacting with the agents of the other users, e.g., for shopping, booking travel, and so on.  This recognition has led to a flurry of industry efforts aimed at standardizing interactions between LLM agents.
Among the prominent ones are the Google-led Universal Commerce Protocol (UCP) \cite{ucp}, which standardizes e-commerce interactions such as checkout and order processing, and the Agent2Agent (A2A) protocol \cite{Google-25:A2A}, which provides a general-purpose framework for interagent task delegation. These protocols are specified informally in terms of JSON schemas, prose descriptions, and example HTTP traces.

\emph{Contributions.} Our objective is to establish the relevance of the ``academic'' declarative approaches for ``practical'' uses.  To this end, we contribute \emph{Strabo}, a general method for exploiting declarative protocols in Langshaw in the context of ongoing industry efforts that model protocols via JSON-RPC or REST API calls.  
The method has two components: (1) Mapping the industry effort into a Langshaw protocol and implementing agents using Peach \cite{AAMAS-26:Peach}, a programming model for Langshaw, and (2) Creating a bridge layer that enables Peach agents to interoperate with industry agents, thus demonstrating interoperation with legacy agents built with conventional methods.

We explain Strabo by modeling UCP's \pname{Checkout} capability as a Langshaw protocol and implementing Peach agents to enact that protocol.
We demonstrate the interoperability of these Peach agents with sample agents available in a UCP code repository\footnote{UCP sample code is available at \url{https://github.com/Universal-Commerce-Protocol/samples}} to establish the fidelity of the Langshaw protocol with respect to UCP.

This exercise establishes that formal, declarative protocols are directly applicable to industry efforts. Whereas UCP is informally specified, leaving important aspects ambiguous, the Langshaw model of UCP specifies interaction constraints formally, bringing precision without sacrificing practical interoperability.  Our exercise also surfaced implicit assumptions in UCP and revealed enhancements required by Langshaw and Peach.  

\section{Background on UCP, Langshaw, and Peach}
We now describe the existing approaches that we build upon in this paper.

\subsection{UCP, the Universal Commerce Protocol}

UCP addresses e-commerce interactions involving the roles of \rname{Platform} (customer), \rname{Business} (merchant), \rname{Credential Provider} (digital wallets), and \rname{Payment Provider} (e.g., PayPal and Stripe).
The idea is that \rname{Platform}, as representative of the shopper, shops with \rname{Business} and uses the \rname{Credential Provider} to generate payment tokens that the \rname{Business} redeems with the \rname{Payment Provider}.  

A typical UCP interaction proceeds as follows.
Both \rname{Platform} and \rname{Business} publish UCP profiles that list their capabilities.
Capabilities are the broad activities involved in an e-commerce interaction, such as \pname{Checkout}, \pname{Order}, and \pname{Identity Linking} (for computing loyalty rewards, and so on).
Capabilities may be extended. For example, \pname{Fulfillment} extends \pname{Checkout} to support the delivery of physical goods.  Based on its own profile and that of \rname{Platform}, \rname{Business} determines if they can interoperate.

\subsubsection{Capabilities.}
A capability is defined by a JSON schema. Table~\ref{tab:checkout-schema} gives some fields in the \pname{Checkout} capability (the full schema is in Appendix~\ref{app:full-checkout}).

\begin{table}[ht]
\centering
\begin{tabular}{l p{2.79cm} c p{5.8cm} }
\toprule
\textbf{Field} & \textbf{Type} & \textbf{Required} & \textbf{Description} \\ \midrule
\paraname{id} & string & Yes & Unique identifier of the checkout session. \\
\paraname{line\_items} & LineItemResponse[] & Yes & List of line items being checked out. \\
\paraname{buyer} & Buyer & No & Representation of the buyer. \\
\paraname{status} & string & Yes & Checkout state indicating the current phase and required action. \\
\paraname{totals} & TotalResponse[] & Yes & Different cart totals. \\
\paraname{messages} & Message[] & No & List of messages with error and info about the checkout session state. \\
\paraname{payment} & PaymentResponse & Yes & Payment configuration and handler info. \\
\paraname{order} & OrderConfirmation & No & Details about an order created for this session. \\
\bottomrule
\end{tabular}
\caption{The \pname{Checkout} capability (extracted from \cite{ucp}).}
\label{tab:checkout-schema}
\end{table}

\subsubsection{Operations.}
All operations on an instance of the \pname{Checkout} capability---i.e., \mname{Create}, \mname{Get}, \mname{Update}, \mname{Complete}, and \mname{Cancel}---are performed by \rname{Platform}.
Each operation specifies an input and an output schema comprising required and optional fields.
The output schema specifies \pname{Checkout} instances that are returned by \rname{Business} in response.
\rname{Business} recomputes the instance based on the input in an \mname{Update}.
Listings~\ref{lst:create} and \ref{lst:create-response}, with minor divergences from UCP documentation, show a \mname{Create} and its (truncated) response.

\begin{lstlisting}[caption={A \mname{Create} operation.},label={lst:create}]
POST /checkout-sessions HTTP/1.1
{"line_items": [
  {"item": {
     "id": "item_123",
     "title": "Red T-Shirt",
     "price": 2500},
   "id": "li_1", "quantity": 2}]}
\end{lstlisting}

\begin{lstlisting}[caption={Response to Listing~\ref{lst:create} (truncated).},label={lst:create-response},mathescape=false]
HTTP/1.1 201 Created
{
  "id": "chk_1234567890",
  "status": "incomplete",
  "messages": [{"type": "error", "code": "missing",
                "path": "$.buyer.email",
                "content": "Buyer email is required"}],
  "currency": "USD",
  "totals": [{"type": "subtotal", "amount": 5000},
             {"type": "tax", "amount": 400},
             {"type": "total", "amount": 5400}],
  "payment": {"instruments": [/*...*/]},
  "line_items": [/*...*/], "links": [/*...*/]
}
\end{lstlisting}

The response indicates that the buyer's email is missing. \rname{Platform} addresses this exception with an \mname{Update} (\texttt{PUT /checkout-sessions/\{id\}}) that supplies the \paraname{buyer} and, optionally, the \paraname{fulfillment} fields; \rname{Business} recomputes and returns the full instance.
Once the instance reaches status \val{ready\_for\_complete}, \rname{Platform} sends \mname{Complete} (\texttt{POST /checkout-sessions/\{id\}/complete}) with payment instrument data.

\subsection{Langshaw}
\label{sec:langshaw}

Langshaw \cite{IJCAI-24:Langshaw} is a declarative protocol language for specifying multiagent interactions. We illustrate its concepts using the checkout protocol in Listing~\ref{lst:atomic-protocol}, which we discuss in detail in Section~\ref{sec:modeling}.

A Langshaw protocol defines \emph{who} interacts (roles), \emph{what} constitutes a complete enactment, and \emph{what they do} (actions with attributes). The \fbf{who} clause names the roles---in our example, \rname{Platform} and \rname{Business}. The \fbf{do} clause lists actions, each performed by a specific role and with some attributes. For instance, \mname{Create} is performed by \rname{Platform} and carries attributes such as \paraname{line\_items} and \paraname{buyer}.

Some attributes are designated as \emph{key}. Keys correlate actions into \emph{enactments}, that is, protocol instances.  A protocol declares one or more attributes as keys in its \fbf{what} clause; all actions bearing the same key values belong to the same enactment. In \pname{SimpleUCP}, \paraname{cid} is the sole key, so all actions sharing a \paraname{cid} value constitute one checkout session.

An attribute can only be bound once within a given enactment; once bound, it is immutable. Performing an action either reuses the binding of each parameter or generates a new one. Action names are themselves bound as attributes when the action occurs, so an action that references another action's name depends on it having occurred. For example, \mname{Created} references \mname{Create}, establishing a causal dependency: \mname{Created} can only occur after \mname{Create}.

The \fbf{sayso} clause assigns data ownership---which role may bind which attributes. In the simplest case, a sayso entry gives a role absolute ownership over an attribute: only that role may produce its binding. However, sayso entries may also specify priority orderings among roles, allowing controlled sharing of attributes where one role's binding takes precedence over another's. The \fbf{nono} clause declares mutual exclusion (e.g., \mname{Completed} and \mname{Cancelled} cannot both occur in the same enactment), and the \fbf{nogo} clause declares directional exclusion (e.g., once \mname{Cancel} occurs, \mname{Complete} is blocked, but not vice versa). The \fbf{what} clause also specifies a completion goal: which actions must occur for an enactment to be considered complete.

\subsection{Peach}

Peach \cite{AAMAS-26:Peach} is a programming model for building agents that participate in Langshaw protocols.
A developer instantiates a \fn{PeachAdapter} (our tooling) by specifying a protocol, the role the agent plays, and an executor that handles communication and the underlying state model. The adapter implements the protocol's semantics: it tracks which actions are feasible given the current state of the enactment, enforces sayso and causal dependencies, and manages key correlation. Developers build agent logic on top of this adapter rather than reimplementing protocol constraints in application code.

The adapter exposes a small API. 
Crucially, \fn{adapter.enabled()} returns the actions the agent may currently perform, given the enactment state. The developer selects an action, binds its attribute values via \fn{FeasibleAction.bind()}, and submits the result via \fn{adapter.attempt()}, which returns a promise. The promise resolves when the action is finalized or rejects if the action fails (e.g., due to a conflicting concurrent action), allowing the developer to handle failures explicitly. To react to other agents' actions, the developer registers observation handlers using the \fn{@adapter.observation} decorator. A \fn{@adapter.on\_completion} handler fires when the protocol's completion goal is satisfied. This API is the same regardless of the executor: Peach currently supports a synchronous executor that connects to a \fn{SyncServer} providing shared state with either instant or batched finalization, and an asynchronous executor based on direct message passing.

A key aspect of Peach is its enablement-based programming model (from BSPL \cite{AAMAS-BSPL-11} and Kiko \cite{AAMAS-23:Kiko}). Rather than requiring the developer to determine which actions are valid after each event---consulting the protocol specification and encoding the operational logic in event handlers---the adapter computes the set of currently feasible actions and presents them directly. The developer (or, in principle, an LLM operating the agent) simply selects from the enabled actions and provides bindings. This shifts the burden of protocol compliance entirely to the adapter: the developer makes domain decisions, not protocol decisions.

Operationally, each adapter computes enabled actions from the protocol and the current state of the enactment, and concurrent attempts within the Peach MAS are reconciled at the SyncServer using its configured finalization policy (instant or batched).
Because correlation is by key, agents engaged in overlapping enactments do not interfere; each enactment's state is independent.
A formal operational semantics is given in~\cite{AAMAS-26:Peach}.

\section{Modeling UCP in Langshaw}
\label{sec:modeling}

There are potentially several ways of capturing UCP's \pname{Checkout} capability in Langshaw.  We focus here on an \emph{atomic} variant that captures the simplest complete checkout: all information is submitted in a single action, exposing the essential data dependencies and completion conditions. An \emph{incremental} variant that decomposes updates into separate actions, making data flow and authority explicit, is given in Appendix~\ref{sec:incremental-protocol}. Our exercise surfaces several assumptions that UCP leaves implicit.

\subsection{A Closer Look at UCP \pname{Checkout}}

We make the following remarks about UCP.

\begin{remarks}
\item\label{Remark:UCP-identifier} Each \pname{Checkout} instance has a unique identifier.  Instances are thought of as \emph{sessions}.  The \mname{Create} does not bear the instance identifier. It is generated in the response.

\item\label{Remark:UCP-atomic} Some fields, e.g., \paraname{id}, are atomic whereas others, e.g., \paraname{line\_items}, are composite. 

\item\label{Remark:UCP-optional} Some fields, e.g., \paraname{line\_items}, are \emph{required} in an instance; others, e.g., \paraname{buyer}, are optional.
An operation's response may additionally specify some of the optional fields, e.g., \paraname{order} in the response to a \mname{Complete}, as required.

\item\label{Remark:UCP-override} Updates can be partial, possibly overriding existing information. 

\item\label{Remark:UCP-request-respose} Operations are conceptually request-response in nature, as is evident from their input-output style specification. 

\item\label{Remark:UCP-sequential} UCP implicitly assumes a synchronous model for operations.
Specifically, \rname{Platform} may issue an operation on an instance only after the previous operation on the instance has returned a response.
Otherwise, \rname{Platform} and \rname{Business} could have incompatible views of the instance.
To see this, suppose \rname{Platform} performs an \mname{Update} on an instance and, before it receives a response, also performs a \mname{Complete} on it.
Now, if \mname{Complete} is processed by \rname{Business} before \mname{Update} (maybe because \mname{Update} is delayed in the network), the \paraname{order} placed may not reflect the intentions of \rname{Platform}.
\end{remarks}

\subsection{UCP in Langshaw}

Listing~\ref{lst:atomic-protocol} gives \pname{SimpleUCP}, a checkout protocol in Langshaw that captures the UCP checkout interactions.
The protocol is \emph{atomic} in the sense that \rname{Platform} submits all checkout information in a single \mname{Create} action; \rname{Business} responds with either \mname{Created} (bearing the server-generated \paraname{id}) or \mname{Failed} (bearing \paraname{reason}, indicating that creation was rejected---e.g., due to missing required fields or a server error).
If the checkout was created successfully, \rname{Platform} may \mname{Complete} or \mname{Cancel} it; a successful \mname{Complete} yields a \mname{Completed} action carrying the \paraname{order} details.

\clearpage

\begin{lstlisting}[language=Langshaw,caption={\pname{SimpleUCP} checkout protocol in Langshaw.},label={lst:atomic-protocol}]
SimpleUCP
who Platform, Business
what cid key, Completed or Failed
do
  Platform: Create(cid,line_items, currency, buyer, payment_pref, discount_codes, fulfillment_pref)
  Business: Created(cid, Create, id, totals, payment, discounts, fulfillment)
  Business: Failed(cid, Create, reason)
  Platform: Complete(cid, Created, id, payment_data, risk_signals)
  Business: Completed(cid, Complete, order)
  Platform: Cancel(cid, Created, id)
sayso
  Platform: line_items, currency, buyer, discount_codes, fulfillment_pref, payment_pref, payment_data, risk_signals
  Business: id, totals, payment, discounts, fulfillment, order, reason
nono
  Completed Failed
  Created Failed
\end{lstlisting}

The \pname{nono} clauses enforce mutual exclusion among outcomes: \mname{Created} and \mname{Failed} cannot both occur in a session (creation either succeeds or fails), and \mname{Completed} and \mname{Failed} cannot both occur (a completed checkout cannot also be failed).
Because there is no \pname{nono} between \mname{Complete} and \mname{Cancel}, \rname{Platform} may emit both concurrently.
How they are resolved depends on the synchronization server configuration: an instant-resolution server processes them in arrival order, so the first received wins; a batch-resolution server may accept both as pending and deliver them together, leaving it to \rname{Business} to decide which to honor and respond to.

Table~\ref{tab:atomic-mapping} shows the mapping from UCP operations to \pname{SimpleUCP} actions.
Notice that UCP's \mname{Update} is realized in \pname{SimpleUCP} via two actions: the \mname{Cancel} of the existing checkout and the \mname{Create} of a new one.

\begin{table}[ht]
\centering
\begin{tabular}{l l l l}
\toprule
\textbf{UCP} & \textbf{HTTP} & \textbf{\pname{SimpleUCP}} & \textbf{Role} \\ \midrule
\mname{Create}  & \texttt{POST /checkout-sessions} & \mname{Create}    & \rname{Platform} \\
                & 201 Created                       & \mname{Created}   & \rname{Business} \\
                & 4xx/5xx                           & \mname{Failed}    & \rname{Business} \\
\mname{Get}     & \texttt{GET /checkout-sessions/\{id\}} & ---          & ---              \\
\mname{Update}  & \texttt{PUT /checkout-sessions/\{id\}} & (\mname{Cancel},\mname{Create})  & \rname{Platform}              \\
\mname{Complete}& \texttt{POST .../complete}        & \mname{Complete}  & \rname{Platform} \\
                & 200 OK                            & \mname{Completed} & \rname{Business} \\
\mname{Cancel}  & \texttt{POST .../cancel}          & \mname{Cancel}    & \rname{Platform} \\
\bottomrule
\end{tabular}
\caption{Mapping from UCP operations to SimpleUCP actions. Get is unused because the adapter tracks protocol state. Update is realized at the protocol level by the Platform agent issuing (Cancel, Create); on the wire, the outbound Create itself is a POST followed by a PUT to supply fields UCP does not accept in the initial request.}
\label{tab:atomic-mapping}
\end{table}


\section{Programming Peach UCP Agents}

We now show how developers build agents for a Langshaw protocol using Peach.
The central point is the separation of concerns: agent code contains only domain decisions (which actions to take and what values to bind), whereas the adapter handles protocol compliance, key correlation, and communication.
We show a \rname{Platform} agent and a \rname{Business} proxy agent for \pname{SimpleUCP}.

\subsection{\rname{Platform} Agent}

The \rname{Platform} agent drives the checkout process. Listing~\ref{lst:platform-setup} shows the adapter setup: the developer loads a protocol, creates an adapter for the \rname{Platform} role, and registers handlers for observations and protocol completion.
Listing~\ref{lst:platform-loop} shows the enablement-driven handlers, each invoked by the adapter when its action becomes feasible.
The adapter also ensures consistency and correctness by presenting only currently valid actions, propagating bindings from prior observations, and accepting via \fn{FeasibleAction.bind()} only attributes the agent has sayso over.

\begin{lstlisting}[caption={\rname{Platform} agent core: adapter setup and observation handlers.},label={lst:platform-setup},language=Python]
protocol = LangshawProtocol.load("protocols/atomic-ucp.lsh")
adapter = PeachAdapter("platform", protocol, "Platform", executor)
done = asyncio.Event()
@adapter.on_completion
async def on_done(keys, mas_id):
    done.set()
@adapter.observation('Created', 'Failed', 'Completed')
async def on_observation(action, performer):
    print(f"Observed {action.action.name} by {performer}")
\end{lstlisting}

\begin{lstlisting}[caption={\rname{Platform} agent enablement-driven handlers: the adapter invokes each handler when its action becomes feasible.},label={lst:platform-loop},language=Python]
@adapter.on_enabled('Create')
async def on_create_enabled(fa):
    await adapter.attempt([fa.bind(
        line_items=SAMPLE_LINE_ITEMS, currency="USD",
        buyer=SAMPLE_BUYER, payment_pref={},
        discount_codes={"codes": []},
        fulfillment_pref=SAMPLE_FULFILLMENT_PREF)])
@adapter.on_enabled('Complete')
async def on_complete_enabled(fa):
    await adapter.attempt([fa.bind(
        payment_data=SAMPLE_PAYMENT_DATA,
        risk_signals={"ip": "127.0.0.1"})])
\end{lstlisting}

\section{Interoperation with Google's UCP Agents}

\subsection{Architecture}

\begin{figure}[ht]
\centering
\begin{tikzpicture}[
    >=Stealth,
    comp/.style={draw,rounded corners=3pt,minimum width=1.5cm,minimum height=0.70cm,
                 font=\small\sffamily,align=center,thick,inner sep=3pt},
    platcomp/.style={comp,fill=figPlatLight,draw=figPlatBlue},
    proxycomp/.style={comp,fill=figProxyLight,draw=figProxyGreen},
    extcomp/.style={comp,fill=figExtLight,draw=figExtGray},
    glabel/.style={font=\scriptsize\sffamily\itshape,text=black!55},
    elabel/.style={font=\scriptsize\sffamily,fill=white,inner sep=1pt},
    biarr/.style={thick},
    harr/.style={thick,draw=orange!70!black},
  ]
  \node[platcomp] (pa)   at (0,0)      {Platform\\Agent};
  \node[platcomp] (sync) at (3,0)    {Sync\\Server};
  \node[proxycomp] (px)  at (6.2,0)    {Proxy\\Adapter};
  \node[extcomp]  (mc)   at (9.7,0)    {UCP Business\\(Google)};
  \begin{scope}[on background layer]
    \node[draw=figPlatBlue!50,dashed,rounded corners=6pt,fill=figPlatLight!40,inner sep=10pt,
          fit=(pa)(px),
          label={[glabel]above:Peach MAS}] {};
    \node[platcomp,
      inner sep=8pt,
      fit=(px),
      align=center,
      label={[font=\small\sffamily,yshift=-12pt]north:Business Proxy}] (bp) {};
    \node[draw=figExtGray!50,dashed,rounded corners=6pt,fill=figExtLight!40,inner sep=8pt,
          fit=(mc),
          label={[glabel]above:External}] {};
  \end{scope}
  \draw[biarr,draw=figPlatBlue]  (pa)   -- node[elabel,above]{\scriptsize SimpleUCP} (sync);
  \draw[biarr,draw=figPlatBlue]  (sync) -- node[elabel,above]{\scriptsize SimpleUCP}  (bp);
  \draw[harr]                    (bp)   -- node[elabel,above]{\scriptsize REST UCP} (mc);
\end{tikzpicture}
\caption{Strabo Architecture. The Peach MAS interoperates with Google's UCP merchant via a Strabo proxy, which translates Langshaw actions in \pname{SimpleUCP} to REST calls using built-in field mapping configured via the \fn{route()} DSL.}
\label{fig:architecture}
\end{figure}
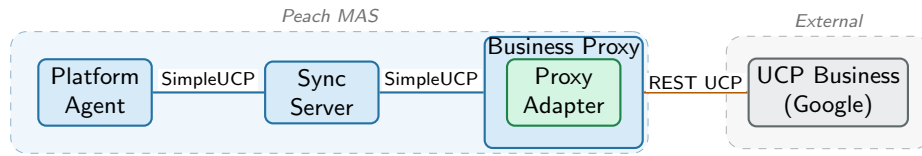

Figure~\ref{fig:architecture} shows the runtime architecture.
The Peach MAS comprises a \rname{Platform} agent that drives the checkout flow and a \rname{Business} proxy adapter that bridges to the external merchant.
Both sides share access to a SyncServer that maintains the protocol's social state.
When the \rname{Platform} agent performs a \pname{SimpleUCP} action, the SyncServer notifies the \fn{Business Proxy}, which uses its built-in field mapping to construct one or more corresponding UCP requests, sends them to the actual UCP \rname{Business} over HTTP, parses the responses, and performs the corresponding \pname{SimpleUCP} action.
Field-name translation is configured inline via the \fn{route()} DSL; neither the \rname{Platform} agent nor the proxy contains hand-written HTTP code, with the only imperative code being orchestration for actions that map to multiple HTTP calls (such as \mname{Create}, which requires a \texttt{POST} followed by a \texttt{PUT}).

\subsection{Proxy}

The proxy bridges the Langshaw protocol to the external merchant HTTP API. \fn{ProxyAdapter} is a \fn{PeachAdapter} subclass with built-in HTTP client and field mapping. Routes are defined via the \fn{route()} DSL, and for each route, the adapter auto-registers an observation handler that fires the HTTP request and submits the response action. For \pname{SimpleUCP}, \mname{Create} requires orchestration: a single Langshaw \mname{Create} maps to a POST to \texttt{/checkout-sessions} followed by a PUT (to supply fields like \paraname{discounts} and \paraname{fulfillment} that UCP does not accept in the initial request). Listing~\ref{lst:business-setup} shows the core of the proxy: routes are declared inline, and the \mname{Create} action is excluded from auto-proxy so that a custom handler can orchestrate the two HTTP calls. All other actions (e.g., \mname{Complete}, \mname{Cancel}) are handled automatically.

\begin{lstlisting}[caption={\fn{ProxyAdapter} setup with \fn{route()} DSL and custom \mname{Create} orchestration.},label={lst:business-setup},language=Python,mathescape=false]
proxy = ProxyAdapter(
    "atomic_proxy", protocol, "Business", executor,
    base_url=merchant_url, internal=["cid"],
    header_hook=ucp_header_hook, exclude=["Create"])
proxy.route("Create -> Created", "POST /checkout-sessions")
proxy.route("Update -> Updated", "PUT /checkout-sessions/{id}",
    request={"discount_codes": "$.discounts",
             "fulfillment_pref": "$.fulfillment",
             "payment_pref": "$.payment"})
proxy.route("Complete -> Completed",
    "POST /checkout-sessions/{id}/complete")
proxy.route("Cancel -> Cancelled",
    "POST /checkout-sessions/{id}/cancel")
@proxy.observation('Create')
async def on_create(action, performer):
    if performer == proxy.name: return
    create_resp = await proxy.request("Create",
        {k: action.bindings[k]
         for k in ("line_items", "currency", "buyer")})
    update_resp = await proxy.request("Update",
        {**action.bindings, "id": create_resp["id"]})
    combined = {**action.bindings, **create_resp, **update_resp}
    await proxy.attempt([proxy.build_action("Created", combined)])
\end{lstlisting}

For each \fn{route()} declaration, \fn{ProxyAdapter} auto-generates an observation handler that fires when the trigger action is observed, sends the corresponding HTTP request, and submits a fixed response action back into the protocol.
This assumes a one-to-one mapping between request and response actions.
Protocols with alternative responses (e.g., \mname{Create} yielding either \mname{Created} or \mname{Failed} depending on the HTTP status) are supported but require a custom observation handler that inspects the response and selects the appropriate action, as illustrated above for \mname{Create}.

\subsection{Field Mapping}

Field mapping is configured inline via the \fn{route()} DSL (visible in Listing~\ref{lst:business-setup}). Each \fn{route()} call declares a round-trip: a pair of Langshaw actions mapped to an HTTP endpoint. The adapter supports three kinds of mappings. \emph{Identity} mappings require no configuration: when a Langshaw attribute name matches the corresponding JSON field name (e.g., \paraname{buyer} $\leftrightarrow$ \texttt{\$.buyer}), the adapter translates automatically. \emph{Explicit} mappings handle cases where names differ: the \fn{request} and \fn{response} keyword arguments specify non-identity translations (e.g., \paraname{discount\_codes} $\leftrightarrow$ \texttt{\$.discounts}, or \paraname{payment\_pref} $\leftrightarrow$ \texttt{\$.payment}). \emph{Excluded} attributes---those declared via \fn{internal}---never appear in HTTP traffic (e.g., \paraname{cid}, which is the Langshaw enactment key).
The complete mapping between UCP fields and \pname{SimpleUCP} attributes is given in Appendix~\ref{sec:field-mapping}.

\subsection{Interactions}
\begin{figure}[ht]
\centering
\begin{tikzpicture}[
    >=Stealth,
    actorbox/.style={draw=#1,fill=#1!12,rounded corners=2pt,thick,
                     minimum width=2.0cm,minimum height=0.6cm,
                     font=\small\sffamily\bfseries,align=center},
    lifeline/.style={draw=#1!50,thick,dashed},
    lmsg/.style={->,thick,draw=figPlatBlue},
    hmsg/.style={->,thick,draw=orange!80!black},
    hmsgrtn/.style={->,thick,dashed,draw=orange!80!black},
    lmsgrtn/.style={->,thick,dashed,draw=figPlatBlue},
    mlabel/.style={font=\scriptsize\sffamily,fill=white,inner sep=1pt},
    dlabel/.style={font=\scriptsize\sffamily\itshape,text=black!50},
  ]
  \def\xP{0} \def\xB{4.2} \def\xM{8.2}
  \node[actorbox=figPlatBlue]  at (\xP,0) (hP) {Platform};
  \node[actorbox=figProxyGreen] at (\xB,0) (hB) {Business Proxy};
  \node[actorbox=figExtGray]   at (\xM,0) (hM) {Business};
  \draw[lifeline=figPlatBlue]  (\xP,-0.35) -- (\xP,-8.0);
  \draw[lifeline=figProxyGreen] (\xB,-0.35) -- (\xB,-8.0);
  \draw[lifeline=figExtGray]   (\xM,-0.35) -- (\xM,-8.0);
  \draw[lmsg]    (\xP,-0.9) -- node[mlabel,above]{\mname{Create}(\paraname{cid},\paraname{line\_items},\paraname{buyer},\ldots)} (\xB,-0.9);
  \draw[hmsg]    (\xB,-1.6) -- node[mlabel,above]{\texttt{POST /checkout-sessions}} (\xM,-1.6);
  \draw[hmsgrtn] (\xM,-2.3) -- node[mlabel,above]{\texttt{201 \{id, status,\ldots\}}} (\xB,-2.3);
  \draw[hmsg]    (\xB,-2.9) -- node[mlabel,above]{\texttt{PUT /checkout-sessions/\{id\}}} (\xM,-2.9);
  \node[font=\scriptsize\sffamily\itshape,text=black!50,anchor=west, text width=54pt] at (\xM+0.15,-2.85) {fulfillment,\\ discounts,\ldots};
  \draw[hmsgrtn] (\xM,-3.5) -- node[mlabel,above]{\texttt{200 \{totals,\ldots\}}} (\xB,-3.5);
  \draw[lmsgrtn] (\xB,-4.2) -- node[mlabel,above]{\mname{Created}(\paraname{cid},\paraname{id},\paraname{totals},\paraname{payment})} (\xP,-4.2);
  \draw[lmsg]    (\xP,-5.1) -- node[mlabel,above]{\mname{Complete}(\paraname{cid},\paraname{id},\paraname{payment\_data})} (\xB,-5.1);
  \draw[hmsg]    (\xB,-5.8) -- node[mlabel,above]{\texttt{POST /checkout-sessions/\{id\}/complete}} (\xM,-5.8);
  \draw[hmsgrtn] (\xM,-6.5) -- node[mlabel,above]{\texttt{200 \{order,\ldots\}}} (\xB,-6.5);
  \draw[lmsgrtn] (\xB,-7.2) -- node[mlabel,above]{\mname{Completed}(\paraname{cid},\paraname{order})} (\xP,-7.2);
  \draw[decorate,decoration={brace,amplitude=4pt,raise=3pt},thick,draw=figPlatBlue!60]
    (\xP-0.2,-7.4) -- (\xP-0.2,-0.7) 
    node[midway,left=7pt,dlabel,rotate=90,anchor=south]{Langshaw};
  \draw[decorate,decoration={brace,amplitude=4pt,raise=3pt,mirror},thick,draw=orange!70]
    (\xM+0.04,-6.7) -- (\xM+0.04,-1.4) 
    node[midway,right=7pt,dlabel,rotate=-90,anchor=south]{HTTP/REST};
\end{tikzpicture}
\caption{Interactions for a \pname{SimpleUCP} checkout via the proxy. \rname{Platform} speaks Langshaw; the proxy bridges to the merchant's HTTP API using its built-in field mapping. The proxy-internal \texttt{PUT} supplies fields (e.g., \paraname{discounts} and \paraname{fulfillment}) that UCP does not accept in the initial \texttt{POST}. Solid arrows: requests; dashed: responses.}
\label{fig:interaction-flow}
\end{figure}
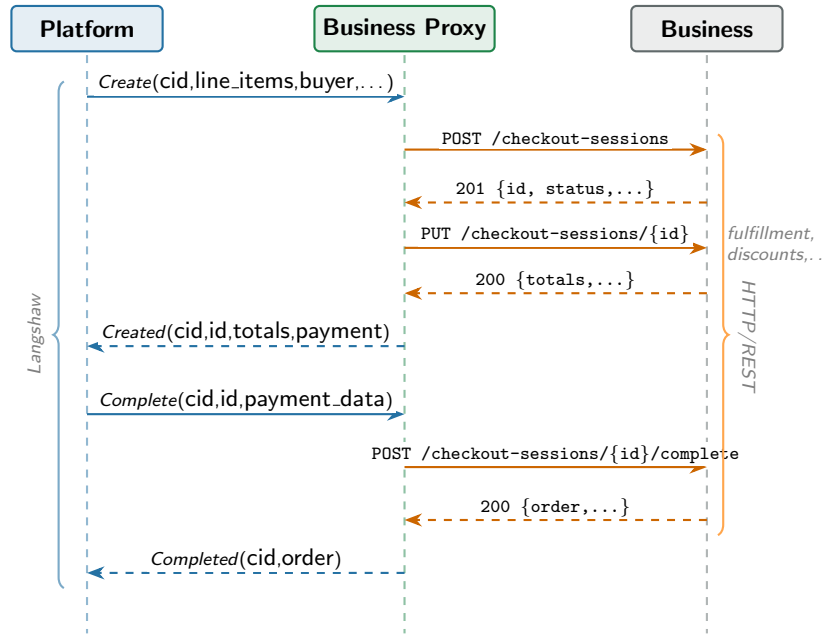

Figure~\ref{fig:interaction-flow} shows the interactions for a complete \pname{SimpleUCP} checkout. Each Langshaw action submitted by \rname{Platform} is intercepted by the proxy, translated into HTTP calls using the route's field mapping, and the response is returned as a Langshaw observation---keeping the \rname{Platform} agent entirely free of HTTP concerns.

\section{Evaluation}
\label{sec:evaluation}

\subsection{Evaluation Criteria}

We evaluate our approach along four dimensions. \emph{Clarity}: Does the Langshaw specification make implicit assumptions explicit? \emph{Generality}: Does the formalism naturally support protocol variants that UCP does not distinguish? \emph{Fidelity}: Can Peach agents interoperate with Google's UCP reference implementation? \emph{Economy}: How much agent code is needed and what fraction is domain logic versus boilerplate?

\subsection{Clarity and Generality}
We focus on how the \pname{SimpleUCP} Langshaw protocol addresses the UCP features noted in the remarks above.

\paragraph{Checkout identity (Remark~\ref{Remark:UCP-identifier}).}
In UCP, checkouts are identified by \paraname{id}, which is generated by \rname{Business} in response to \rname{Platform}'s \mname{Create}.  Such a model not only reflects a centralized mindset, but also is incompatible with meaning: Being unidentified, a UCP \mname{Create} by itself cannot be meaningful.  Since Langshaw's purpose is to support meaning, it identifies every action instance.  In \pname{SimpleUCP}, this is accomplished by declaring \paraname{cid} as the key for every action, including \mname{Create}.  For compatibility with UCP, \pname{SimpleUCP} introduces the (non-key) attribute \paraname{id}, over which \rname{Business} has sayso; it is generated by \rname{Business} in \mname{Created}, and is used by \rname{Platform} in subsequent actions such as \mname{Complete} and \mname{Cancel}.  For interoperation with Google's sample agents, the proxies strip out \paraname{cid}.

\paragraph{Atomic versus composite fields (Remark~\ref{Remark:UCP-atomic}).}
UCP distinguishes atomic fields (e.g., \paraname{id}, a string) from composite ones (e.g., \paraname{line\_items}, an array of objects). Langshaw treats all attributes uniformly; their internal structure is orthogonal to the protocol specification. Peach handles atomic and composite values equally well by treating composite fields as opaque values compared using deep equality (a built-in Python feature over dictionaries) and serializing all values for storage as JSON, which is compatible with the format received from UCP.

\paragraph{Modeling optional attributes (Remark~\ref{Remark:UCP-optional}).}
UCP marks certain fields as optional: \paraname{buyer}, \paraname{fulfillment}, and \paraname{discount\_codes} may or may not be supplied.  One way to capture such optionality is via null values.  In the \pname{SimpleUCP} protocol, all attributes appear on the \mname{Create} action.  An attribute that the agent does not wish to supply is bound to a null value; the proxy strips null-valued fields before constructing the HTTP request. This approach is concise but relies on a convention external to the protocol specification: Langshaw does not distinguish a deliberately null binding from a missing one.  A better approach would be to add support for \texttt{optional} action attributes in the Langshaw language.  We leave this extension as future work.  

The incremental variant in Appendix~\ref{sec:incremental-protocol} avoids null values altogether by giving each optional concern its own action.

\paragraph{Overwrite semantics (Remark~\ref{Remark:UCP-override}).}
UCP allows \rname{Platform} to selectively update fields that have already been bound in a session.
In Langshaw, attributes may be bound at most once in an enactment; they are immutable thereafter.
Moreover, \pname{SimpleUCP} models a checkout as a complete, immutable object.
To revise a checkout, \rname{Platform} cancels the current session and creates a new one with the updated data---a cancel-and-recreate pattern that is semantically equivalent to an in-place update.
This comes at some cost: a \mname{Cancel} is ambiguous, as observers cannot distinguish a genuine user cancellation from a revision.
In exchange, each session has a definitive state, and it is always clear which version of the checkout is being completed.
Mutable amendment would conflict with Langshaw's commitment to immutability, which lets agents reason about enactment state without a shared mutable store.
The incremental variant's version key (Appendix~\ref{sec:incremental-protocol}) is the principled counterpart to UCP's in-place update: each amendment is a fresh subenactment, and the application interprets the collection of versions as its domain requires (e.g., treating the highest sortable \paraname{v} as authoritative).
The incremental versioning example is deliberately minimal; a richer structure---e.g., recording parent--child lineage or scoping versioning to attribute groups---could more closely reflect UCP's selective-update semantics without relaxing immutability.

\paragraph{Request-response and synchronous ordering (Remarks~\ref{Remark:UCP-request-respose} and~\ref{Remark:UCP-sequential}).}
UCP operations are conceptually request-response pairs, and UCP implicitly assumes that operations on a session are issued sequentially. Langshaw replaces both conventions with structural causal dependencies: an action that references another action's name can only occur after it. \pname{SimpleUCP} is inherently synchronous---a single \mname{Create}--\mname{Complete} exchange---so no additional ordering mechanism is needed. The incremental variant makes the design choice explicit by declaring \paraname{v} (version) as key (Appendix~\ref{sec:incremental-protocol}). Attribute binding authority, implicit in UCP's JSON schemas, is made explicit through \fbf{sayso}.

UCP additionally requires idempotency keys in HTTP headers to prevent the server from reprocessing a duplicate request and repeating side-effects (e.g., creating a second order).
The UCP sample generates a fresh random key per request (Listing~\ref{lst:sdk-create}), which provides protection only at the network level---where the same request is resent with its original headers intact---but not against application-level retries where a new key would be generated.
In our implementation, idempotency key generation is encapsulated in the \fn{ProxyAdapter}, keeping this transport concern transparent to the agent.

Table~\ref{tab:remarks-mapping} summarizes how each remark is addressed.

\begin{table}[ht]
\centering
\begin{tabular}{c p{5.5cm} p{5.5cm}}
\toprule
\textbf{\#} & \textbf{UCP Model} & \textbf{\pname{SimpleUCP} Model} \\ \midrule
1 & \mname{Create} lacks instance ID & Explicit key \paraname{cid}; \mname{Created} introduces \paraname{id} \\
2 & Atomic vs.\ composite fields & Uniform attributes; types orthogonal \\
3 & Required vs.\ optional fields & Null values (\pname{SimpleUCP}) or separate actions (incremental) \\
4 & Overwrite semantics & Avoided in \pname{SimpleUCP}; version key in incremental (Appendix) \\
5 & Request-response convention & Structural causal dependencies \\
6 & Synchronous ordering (implicit) & \pname{SimpleUCP} is inherently synchronous; incremental uses \paraname{v} key (Appendix) \\
\bottomrule
\end{tabular}
\caption{How Langshaw addresses assumptions implicit in UCP's specification.}
\label{tab:remarks-mapping}
\end{table}

\subsection{Fidelity: Interoperation with Google's UCP Agents}

We tested interoperation end-to-end by connecting a Peach \rname{Platform} agent to Google's UCP reference \rname{Business} server via the \fn{ProxyAdapter}.
The \rname{Platform} agent successfully completed checkout sessions with the Google server, with the adapter's built-in field mapping handling name translation and session ID correlation transparently.

\subsection{Economy: Comparison with Google's UCP Agent}

We compare the Peach agent implementation with a comparable agent built directly against Google's UCP SDK. In the Peach agent, ordering constraints, session management, and data ownership are enforced by the adapter---the agent code contains no ordering logic. In the direct HTTP agent, the developer must manually track session identifiers, check status before completing (e.g., verifying \val{ready\_for\_complete}), and construct correct HTTP requests. The \rname{Business} proxy is particularly compact: the \fn{ProxyAdapter} auto-generates observation handlers from \fn{route()} declarations, so adding a new UCP capability (e.g., Orders) requires only a new protocol file and a few route definitions, not new agent code.

Table~\ref{tab:loc-comparison} compares lines of code across implementations. The difference is most visible in equivalent operations. Listings~\ref{lst:sdk-create} and \ref{lst:peach-create} show checkout creation in the direct HTTP client and the Peach agent, respectively.

\begin{lstlisting}[caption={Checkout creation using the UCP SDK directly.},label={lst:sdk-create},language=Python]
item1 = ItemCreateRequest(id="bouquet_roses", title="Red Rose")
line_item1 = LineItemCreateRequest(quantity=1, item=item1)
payment_req = PaymentCreateRequest(
    instruments=[], selected_instrument_id=None,
    handlers=supported_handlers)
buyer_req = Buyer(full_name="John Doe", email="john@example.com")
create_payload = CheckoutCreateRequest(
    currency="USD", line_items=[line_item1],
    payment=payment_req, buyer=buyer_req)
headers = get_headers()  # UCP-Agent, signature, idempotency-key
json_body = create_payload.model_dump(
    mode="json", by_alias=True, exclude_none=True)
response = client.post(
    "/checkout-sessions", json=json_body, headers=headers)
checkout_id = response.json().get("id")
\end{lstlisting}

\begin{lstlisting}[caption={Checkout creation using the Peach adapter.},label={lst:peach-create},language=Python]
@adapter.on_enabled('Create')
async def on_create_enabled(fa):
    await adapter.attempt([fa.bind(
        line_items=SAMPLE_LINE_ITEMS, currency="USD",
        buyer=SAMPLE_BUYER, payment_pref={})])
\end{lstlisting}

\begin{table}[ht]
\centering
\begin{tabular}{l c ccc}
\toprule
 & \textbf{Google} & \multicolumn{3}{c}{\textbf{Peach}} \\
\cmidrule(lr){2-2}\cmidrule(lr){3-5}
\textbf{Component} & REST Client & \pname{SimpleUCP} & Incremental & Direct \\ \midrule
\rname{Platform} agent & 915 & 97 & 113 & 128 \\
\rname{Business} proxy         & ---        & 163 & 171 & 134 \\
\bottomrule
\end{tabular}
\caption{Lines of code comparison for the Checkout capability, with our Peach agent as \rname{Platform} and the Google sample as \rname{Business}. The \rname{Business} proxy includes inline \fn{route()} declarations for field mapping.}
\label{tab:loc-comparison}
\end{table}

\section{Discussion}\label{sec:discussion}

Agent2Agent (A2A) \cite{Google-25:A2A}, originally developed by Google and now managed by the Linux Foundation, is a widely adopted protocol for interagent task delegation. A client discovers remote agents via \emph{AgentCard}s and delegates tasks; tasks progress through states (\val{submitted}, \val{working}, \val{completed}, \val{failed}, \val{canceled}) and may involve multiturn conversations. A2A and UCP are complementary: A2A handles discovery and delegation while UCP handles domain-specific interactions such as checkout. Like UCP, A2A is mostly informally specified; its state machine is described in prose, and constraints such as whether a canceled task may be resumed are ambiguous. A2A's \emph{context identifier} is analogous to a Langshaw enactment key, and the bridge architecture presented here could equally apply: model the A2A task lifecycle as a Langshaw protocol and bridge to A2A's JSON-RPC transport. 

Baldoni \etal~\cite{Baldoni+25:BSPL+SARL} recently demonstrated BSPL integration with the SARL agent language, exploring a different point in the implementation design space.

Chopra et al.~\cite{EUMAS-26:toolkit-chapter} survey the IOP toolkit Strabo builds on---protocol languages, verifiers, and programming models. Strabo's contribution is not in expressiveness or execution semantics, which are inherited from Langshaw and Peach, but in the bridge architecture: the \fn{ProxyAdapter} and \fn{route()} DSL map declarative actions onto informally-specified HTTP APIs, letting new capabilities be added by writing a protocol and a few routes rather than new agent code.

Strabo is concerned with bridging formal protocols to existing HTTP APIs. Chopra and Singh \cite{chopra:fluid:2025} recently proposed Fluid, which brings social norms to web-based multiagent systems---a complementary direction that could inform how normative constraints are layered atop protocols like those modeled here.

Historically, agent communication was standardized through FIPA-ACL \cite{FIPA-02:ACL} and KQML \cite{Finin+94}, which focused on speech-act semantics for individual messages. UCP and A2A represent a new generation of industry protocols that emphasize structured interactions, but share with their predecessors the challenges of informal specification and inadequate support for meaning \cite{Computer-98}.

The end-to-end test demonstrates feasibility but is not an exhaustive evaluation of protocol conformance.

\section{Conclusion}

We have shown that formal, declarative MAS protocol techniques are directly applicable to industry e-commerce protocols. Modeling UCP's Checkout capability in Langshaw made explicit several assumptions---most notably around attribute binding authority and ordering---that UCP's informal specification leaves implicit. The Langshaw formalism naturally supports protocol variants that UCP does not distinguish, exposing a design space that practitioners can reason about; an incremental variant is explored in Appendix~\ref{sec:incremental-protocol}. Peach agents built against the Langshaw protocol interoperate with Google's UCP reference implementation via Strabo's \fn{ProxyAdapter}, which bridges protocol actions to HTTP round-trips with inline field mapping, demonstrating that formal specification and practical interoperability are not at odds. Industry protocols like UCP and A2A are defining the interaction patterns for AI-assisted commerce and multiagent coordination; our exercise demonstrates that MAS protocol research can contribute precision and verifiability to these efforts.

This exercise was focused on operations in the UCP \pname{Checkout} capability; UCP additionally includes \pname{Order}, \pname{Identity Linking}, and extensible capabilities, each of which would need its own protocol and route definitions.
We model only the \rname{Platform}--\rname{Business} interactions; the protocol also involves roles corresponding to credential and payment providers.
Extending Strabo to address additional capabilities and roles should be a rewarding direction, potentially testing Langshaw's support for protocol composition (a core facility in BSPL) and dynamic role bindings (supported in Splee \cite{AAMAS-17:Splee}, a BSPL extension).

Several directions remain for future work.
Adding \texttt{optional} attribute support to Langshaw would address the null-value convention that \pname{SimpleUCP} currently relies on, making optionality a first-class protocol concept rather than an implementation convention.
The bridge architecture generalizes beyond UCP: A2A's task lifecycle could be modeled as a Langshaw protocol and bridged to its JSON-RPC transport, testing the approach against a complementary industry standard with a different interaction style.
Finally, a benefit of Langshaw that we did not exploit is decentralized enactments---where each agent maintains its own local view of the protocol state without a shared server.
Exploiting this property would enable peer-to-peer agent interaction more naturally aligned with the distributed nature of agentic AI deployments, and would test whether the \fn{ProxyAdapter} architecture extends to settings without central coordination.

\emph{Reproducibility.}  Our software is available at \url{https://gitlab.com/masr/strabo}.

\DeclareRobustCommand{\nUmErAL}[1]{#1}\DeclareRobustCommand{\nAmE}[3]{#3}\DeclareRobustCommand{\nUmErAL}[1]{#1}\DeclareRobustCommand{\nAmE}[3]{#3}


\clearpage
\appendix

\section{\pname{Checkout} Capability}
\label{app:full-checkout}

Table~\ref{tab:full-checkout} gives the complete specification of the \pname{Checkout} capability.

\begin{table}[htb!]
\centering
\begin{tabular}{l p{2.79cm} c p{5.8cm} }
\toprule
\textbf{Field} & \textbf{Type} & \textbf{Required} & \textbf{Description} \\ \midrule
\paraname{ucp} & ResponseCheckout & Yes & Protocol metadata and checkout response root. \\
\paraname{id} & string & Yes & Unique identifier of the checkout session. \\
\paraname{line\_items} & LineItemResponse[] & Yes & List of line items being checked out. \\
\paraname{buyer} & Buyer & No & Representation of the buyer. \\
\paraname{status} & string & Yes & Checkout state indicating the current phase and required action. \\
\paraname{currency} & string & Yes & ISO 4217 currency code. \\
\paraname{totals} & TotalResponse[] & Yes & Different cart totals. \\
\paraname{messages} & Message[] & No & List of messages with error and info about the checkout session state. \\
\paraname{links} & Link[] & Yes & Links to be displayed by the \rname{Platform} (e.g., Privacy Policy, TOS). \\
\paraname{expires\_at} & string & No & RFC 3339 expiry timestamp; default TTL often {$\sim$}6 hours. \\
\paraname{continue\_url} & string & No & URL for checkout handoff and session recovery. \\
\paraname{payment} & PaymentResponse & Yes & Payment configuration and handler info. \\
\paraname{order} & OrderConfirmation & No & Details about an order created for this session. \\
\bottomrule
\end{tabular}
\caption{The \pname{Checkout} capability (extracted from \cite{ucp}).}
\label{tab:full-checkout}
\end{table}
\section{Mapping Between UCP Fields and \pname{SimpleUCP} Attributes}
\label{sec:field-mapping}

Table~\ref{tab:field-mapping} gives the complete field-to-attribute mapping used by the \fn{ProxyAdapter}.
Identity mappings use the same name on both sides.
Explicit mappings arise where UCP and \pname{SimpleUCP} use different names for the same data.
Fields marked \emph{dropped} have no protocol equivalent in \pname{SimpleUCP} because their information is either implicit in the action sequence or unnecessary for the Langshaw model.

\begin{table}[htb!]
\centering
\begin{tabular}{l l l}
\toprule
\textbf{UCP Field} & \textbf{\pname{SimpleUCP} Attribute} & \textbf{Mapping} \\
\midrule
\multicolumn{3}{l}{\textit{Checkout response fields}} \\
\paraname{ucp}          & ---                    & dropped (protocol metadata) \\
\paraname{id}           & \paraname{id}          & identity \\
\paraname{line\_items}  & \paraname{line\_items} & identity \\
\paraname{buyer}        & \paraname{buyer}       & identity \\
\paraname{status}       & ---                    & dropped (implicit in action sequence) \\
\paraname{currency}     & \paraname{currency}    & identity \\
\paraname{totals}       & \paraname{totals}      & identity \\
\paraname{messages}     & \paraname{reason}      & partial (error messages $\to$ \mname{Failed}) \\
\paraname{links}        & ---                    & dropped \\
\paraname{expires\_at}  & ---                    & dropped \\
\paraname{continue\_url}& ---                    & dropped \\
\paraname{payment}      & \paraname{payment}     & identity (in \mname{Created}) \\
\paraname{order}        & \paraname{order}       & identity (in \mname{Completed}) \\
\midrule
\multicolumn{3}{l}{\textit{Request fields (sent in \mname{Update}/\mname{Complete})}} \\
\paraname{discounts}    & \paraname{discount\_codes} & explicit (\fn{route()} DSL) \\
\paraname{fulfillment}  & \paraname{fulfillment\_pref} & explicit (\fn{route()} DSL) \\
\paraname{payment}      & \paraname{payment\_pref}   & explicit (\fn{route()} DSL) \\
\paraname{payment\_data}& \paraname{payment\_data}   & identity \\
\paraname{risk\_signals}& \paraname{risk\_signals}   & identity \\
\midrule
\multicolumn{3}{l}{\textit{Langshaw-internal (not in UCP)}} \\
---                     & \paraname{cid}         & enactment key; stripped by proxy \\
\bottomrule
\end{tabular}
\caption{Mapping between UCP fields and \pname{SimpleUCP} attributes.
UCP's \paraname{payment} field is overloaded: in request context it carries payment preferences (\paraname{payment\_pref}); in response context it carries payment configuration (\paraname{payment}).}
\label{tab:field-mapping}
\end{table}

\section{Incremental \pname{Checkout} Protocol}\label{sec:incremental-protocol}

The incremental protocol (Listing~\ref{lst:incremental-protocol}) generalizes \pname{SimpleUCP} to support UCP's multistep update flow.
Rather than submitting all information in a single \mname{Create}, \rname{Platform} may issue separate actions for each concern: \mname{SetBuyer}, \mname{ApplyDiscounts}, \mname{SetFulfillment}, and \mname{SetPayment}, each carrying only its own attributes.
\rname{Business} responds to each with a corresponding acknowledgment.
This decomposition makes data flow explicit---each action pair affects only the attributes it names.
Regarding optional attributes (Remark~\ref{Remark:UCP-optional}): each concern has its own action, so the agent simply omits any action it does not need.
No null values are required; the protocol specification itself captures which combinations of actions are valid.

The incremental protocol introduces a second key, \paraname{v}, which serves as a version identifier.
Each mutation action and its response carry \paraname{v}, and since an attribute may only be bound once per composite key, successive mutations are distinguished by their \paraname{v} values---each operates within its own subenactment.
This also addresses overwrite semantics (Remark~\ref{Remark:UCP-override}): successive updates produce distinct bindings rather than overwriting previous ones, and agents can observe all bindings across subenactments to merge values and selectively override at the application level.
The version key does not itself impose ordering; however, the developer may choose sortable values (e.g., ULIDs) and operate on the latest version, achieving the sequential behavior that UCP assumes by convention (Remark~\ref{Remark:UCP-sequential}).
The protocol makes this design choice explicit.

\begin{lstlisting}[caption={Incremental UCP checkout protocol in Langshaw.}, label={lst:incremental-protocol}, language=Langshaw]
IncrementalUCP
who Platform, Business
what eid key, v key, Completed or Cancelled
do
  Platform: Create(eid, line_items, currency, buyer, payment)
  Business: Created(eid, Create, id, payment_config, messages)
  Business: StatusChange(eid, Created, status, totals, messages)
  Platform: Update(eid, v, Created, id, line_items, currency)
  Business: Updated(eid, v, Update, line_items, messages)
  Platform: SetBuyer(eid, v, Created, id, buyer)
  Business: BuyerSet(eid, v, SetBuyer, buyer, messages)
  Platform: ApplyDiscounts(eid, v, Created, id, discount_codes)
  Business: DiscountsApplied(eid, v, ApplyDiscounts, discounts_applied, messages)
  Platform: SetFulfillment(eid, v, Created, id, fulfillment_pref)
  Business: FulfillmentSet(eid, v, SetFulfillment, fulfillment, messages)
  Platform: SetPayment(eid, v, Created, id, payment_pref)
  Business: PaymentSet(eid, v, SetPayment, payment_config, messages)
  Platform: Complete(eid, v, Create, id, payment_data, risk_signals)
  Business: Completed(eid, v, Complete, order, messages)
  Platform: Cancel(eid, v, Created, id)
  Business: Cancelled(eid, v, Cancel, messages)
  Business: Shipped(eid, Completed, order_id, tracking_number, tracking_url, carrier)
  Business: Delivered(eid, Shipped, order_id)
sayso
  Platform: line_items, currency, buyer, discount_codes, fulfillment_pref, payment_pref, payment_data, risk_signals, payment
  Business: id, status, totals, discounts_applied, fulfillment, payment_config, order, order_id, tracking_number, tracking_url, carrier, messages
nono
  Completed Cancelled
nogo
  Cancel -/> Complete
\end{lstlisting}

\end{document}